%% file: root.tex
\documentclass[letterpaper, 10 pt, conference]{ieeeconf}  

\IEEEoverridecommandlockouts                              

\usepackage{amsmath} 
\usepackage{amssymb}  
\usepackage{graphicx}
\usepackage{capt-of}   
\usepackage{cuted} 
\usepackage{makecell}

\usepackage[hidelinks]{hyperref}

\usepackage{tikz}
\usepackage{pgfplots}
\usepackage{pifont}
\pgfplotsset{compat=1.18}
\usepackage{algorithm}
\usepackage{algpseudocode} 
\floatname{algorithm}{Protocol}
\usepackage{booktabs}
\usepackage{tabularx}
\usepackage{array}

\title{\LARGE \bf
From Wizard-of-Oz Human–Robot Dialogue Collection to a Taxonomy of Robot Response Decisions: A Retrospective Analysis of Assistive Pilot Interactions\vspace{-0.9em}
}

\author{Guangping Liu$^{1}$,
Nicholas Hawkins$^{1}$,
Tipu Sultan$^{1}$,
Flavio Esposito$^{2}$,
Madi Dian$^{1}$%
\thanks{$^{1}$Department of Aerospace and Mechanical Engineering,
$^{2}$Department of Computer Science, Saint Louis University,
St. Louis, MO 63103, United States.
Dr. Madi Dian is the corresponding author
(\texttt{madi.babaiasl@slu.edu}).}%
}

\begin{document}

\maketitle
\begin{strip}
\centering
\vspace{-7em}
\includegraphics[width=\textwidth]{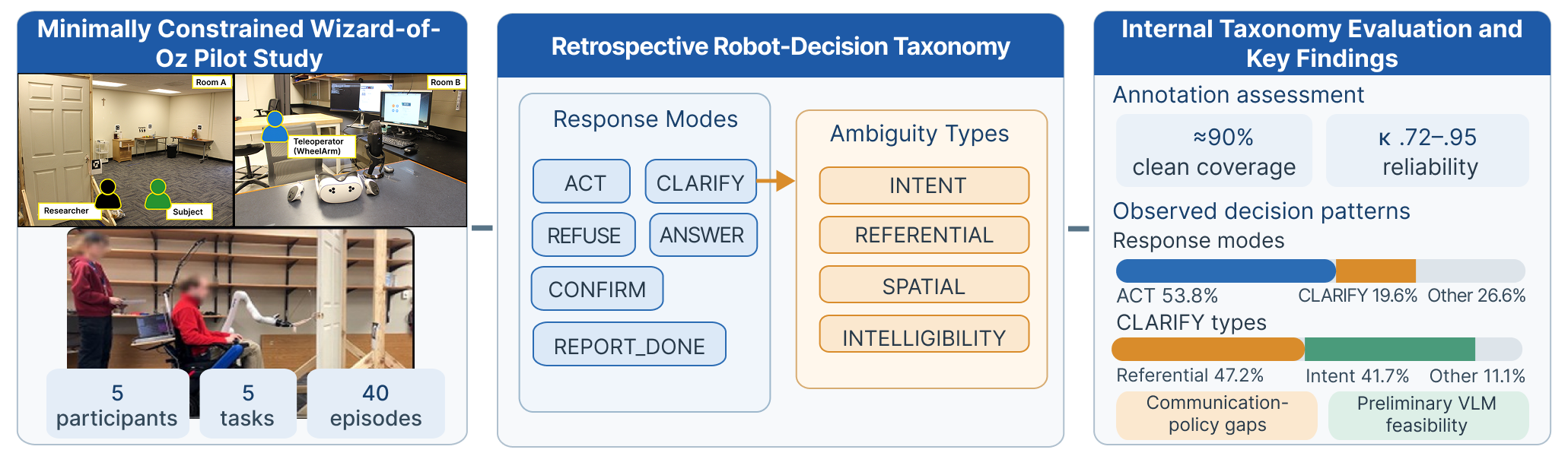}
\vspace{-2em}
\captionof{figure}{\textbf{Overview of the prior WoZ pilot study~\cite{liu2026multimodal} and the present retrospective analysis: Unconstrained Wizard-of-Oz (WoZ) Human-Robot Interaction data collection~\cite{liu2026multimodal} is conducted to collect authentic dialogues and robot data, then the dialogues are retrospectively analyzed to identify robot decision points and develop a hierarchical taxonomy of response modes and clarification-related ambiguity types. The taxonomy is evaluated through coverage and Inter-Rater Reliability (IRR) analyses, then used for the feasibility test of the offline VLM fine-tuning.}}
\label{fig:overview}
\vspace{-1.3em}
\end{strip}
\section{Introduction}
\label{sec:introduction}\input{Sections/Introduction}





\section{Methods}
\label{sec:Methods}\input{Sections/Methods}

\section{Results}
\label{sec:Results}\input{Sections/Results}

\section{Conclusion and Future Work}
\label{sec:Discussion and Future Work}\input{Sections/Conclusion}

\maketitle
\thispagestyle{empty}
\pagestyle{empty} 
\addtolength{\textheight}{-12cm}   

\section{Acknowledgment}
Panels (a) and (b) of Fig.~\ref{fig: llava_finetuned} are reproduced from the arXiv version of the prior pilot work~\cite{liu2026multimodal} under the \href{https://creativecommons.org/licenses/by-nc-sa/4.0/}{CC BY-NC-SA 4.0 license}. The panels were resized and incorporated into the composite figure; their content was otherwise unchanged.




\bibliographystyle{IEEEtran}
\bibliography{reference}

\end{document}

%% file: Sections/Introduction.tex


Despite the potential of natural language interfaces, robots operating in everyday indoor environments must interpret human instructions, which are often incomplete~\cite{chen2023asking}. Commanding robots in practice remains hindered by the inherent ambiguity and vagueness of human dialogue. Such ambiguity typically arises from shared prior knowledge, latent user preferences, and social conventions~\cite{pezzelle2023dealing}, reflecting contextual nuances that are often transparent to humans but inaccessible to robotic systems. Robot clarification therefore requires explicit criteria for deciding when to clarify, act, refuse, or otherwise respond, as well as what information to request. However, few existing datasets capture real-world, task-grounded Human-Robot Dialogue (HRD) together with visual observations and robot behavior to demonstrate how robot can clarify ambiguities. This scarcity limits the empirical understanding of what communication policy assistive robots can use to solve vague instructions. Consequently, robot-side conversational interaction behavior guidance remain underexplored, hindering the downstream conversational behavior learning for assistive HRI systems.

To investigate this gap, previous work~\cite{liu2026multimodal} developed a multimodal Wizard-of-Oz (WoZ) framework and conducted a preliminary study with one teleoperator and five participants. The teleoperator received training in task execution but no formal communication policies, allowing minimally constrained robot-side behavior and unscripted participant responses. The framework collected 53 successful episodes containing synchronized dialogue, visual observations, and robot states. Of these episodes, 40 episodes meeting the dialogue analysis inclusion criteria were retained for retrospective analysis. In the present study, we analyzed these 40 episodes and derived a hierarchical taxonomy of six robot response modes and four clarification-related ambiguity types. Coverage analysis produced clean rates of approximately 89\% to 91\%, while human-human and human-AI comparisons achieved Cohen's $\kappa$ values ranging from 0.72 to 0.95 across the evaluated annotation levels. A preliminary VLM fine-tuning experiment further indicated that the taxonomy-derived labels can provide learnable supervision.

This pilot work connects authentic assistive HRD collection with systematic communication policy and model development, providing an empirical foundation for clarification-aware assistive robots. With retrospective taxonomy validation and modification, this integrated approach has potential to provide a unified methodology for collecting natural dialogue, developing consistent robot communication policies, and training offline conversational models from the same real-world interactions.

%% file: Sections/Methods.tex
\subsection{Wizard-of-Oz Assistive Interaction Framework}
WoZ is an experimental paradigm in which participants interact with a system they believe to be autonomous, while its behaviors are controlled by a human operator \cite{riek2012wizard}. The prior study employed a two-room WoZ setup to simulate the dialogue-based autonomy of WheelArm, an assistive mobile manipulator, and elicit natural participant responses. The teleoperator controlled WheelArm from Room B using a VR-based system extended from OpenTeach \cite{iyer2024open}, while the participant and an accompanying researcher remained in Room A. Physical separation concealed the teleoperator, and participants were informed that WheelArm was capable of natural-language, multi-turn dialogue. Communication was routed through a voice channel, with the teleoperator’s speech converted in real time to maintain a consistent robot persona. During each interaction, the framework synchronized conversational audio, RGB-D observations, IMU measurements, end-effector poses, and whole-body joint states.

\subsection{Task and Dialogue Collection Protocol}
The prior study selected five common indoor assistive tasks, which are feeding, drinking, cleaning, door opening, and drawer opening, based on the International Classification of Functioning, Disability and Health framework \cite{Ustun2003ICF} and practical examples of wheelchair-mounted robotic-arm use. The task configurations included multiple candidate objects, out-of-view targets, and combined navigation and manipulation requirements, creating opportunities for various ambiguity. Data were collected from five participants, with 2 female and 3 male, and none of them had prior assistive technology experience. 

Before data collection, the teleoperator received more than 30 hours of task-execution and safety training. During interaction, the teleoperator observed the environment only through WheelArm's ego and wrist cameras, approximating the robot's partial observability. Participants were introduced to WheelArm as an autonomous assistive robot and completed the tasks through natural, multi-turn verbal interaction without predefined voice commands. To avoid imposing a predetermined communication structure, the teleoperator was not provided with formal communication policy or a clarification taxonomy. Instead, the teleoperator interpreted each instruction using the available perceptual context and requested additional information when the instruction was insufficient for task execution. 


After completing the interactions, participants were debriefed about the WoZ procedure and provided re-consent with the option to withdraw their data. They then completed a brief manual-control comparison and a post-study questionnaire assessing enjoyment of the dialogue-driven interaction, perceived autonomy and preference relative to manual control, and open-ended feedback on the interaction.

In data processing, the past study first filtered the raw episodes to retain only successful teleoperation runs, defined as completing the task without
drops or excessive force on non-target objects. An episode is
labeled successful only if the task is completed without
object drops, items falling, environmental collisions, or inap-
propriate force. The successful teleoperation episodes are then filtered by the dialogue inclusion criteria, which are only conversations that are complete and usable from both speakers, excluding those with missing context or responses before execution, teleoperation system-related dialogue, and audio failures.

\subsection{Retrospective Dialogue Analysis and Taxonomy}
With the filtered data, we annotated only robot (wizard) turns using a hierarchical scheme. First, a robot turn was marked as a decision point when it responds to a new instruction, a change in the task goal, object, or destination, or the completion of an action; motion corrections, same-target refinements, greetings, meta-speech, and sensor-only rows were excluded. Each decision point was then assigned one mutually exclusive response mode: ANSWER for responding to a user question or providing reassurance, REPORT\_DONE for announcing task completion, REFUSE for impossible or safety-limited requests, CONFIRM for verifying a single task inferred by the robot, CLARIFY when essential task information remained unresolved, and ACT when the action, object, and required destination were sufficiently specified. Turns labeled CLARIFY were additionally categorized by the missing information: intent for an unspecified goal or action, referential for multiple visible candidate objects, spatial for an unknown location or out-of-view target, and intelligibility for speech that could not be reliably interpreted. 

\subsection{Taxonomy Coverage and Reliability Evaluation}
Two human annotators and an AI annotator independently applied the proposed annotation scheme to the pilot dataset. Reliability is evaluated using decision-point detection, robot response mode, and ambiguity type. We report observed agreement and Cohen's $\kappa$ for each target. To assess annotation confidence and taxonomy coverage, two human annotators categorize each label as clean, overlapping, forced, or uncovered, indicating a clear fit, multiple plausible labels, an imperfect but closest available category, or no suitable category, respectively.

\subsection{Feasibility Test of Taxonomy-Label Learnability}
We evaluated the feasibility of whether taxonomy-derived labels can supervise an offline VLM using ego- and wrist-camera images, dialogue history, and the current user instruction as inputs, with robot mode, ambiguity type, and response text as targets. To prevent interaction leakage, complete episodes were assigned to train, validation, or test sets, with ambiguity-focused augmentation applied only to training data. Ambiguity-focused augmentation was applied only to training episodes, while validation and test sets remained real data only. We selected only ACT and CLARIFY (intent, referential, and spatial) for training because these modes are more readily balanced through augmentation. We then computed confusion matrices for response modes and ambiguity types to assess the learnability of the taxonomy-derived annotations.

%% file: Sections/Results.tex
\begin{figure*}
    \centering   
    \includegraphics[width=\linewidth]{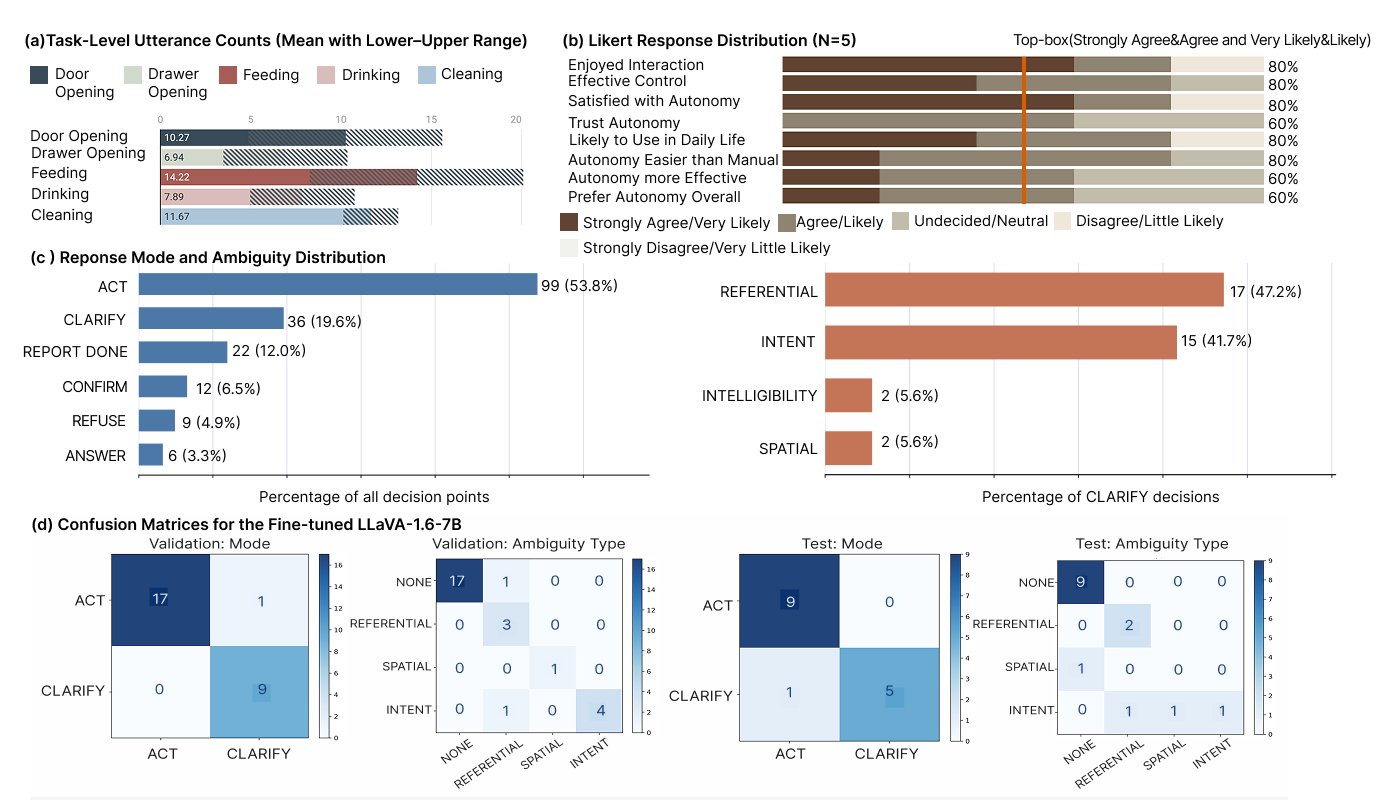}
    \vspace{-2em}
    \caption{(a) Utterance counts by task. (b) Likert-response distributions indicating positive perceptions of dialogue-based assistive interaction. (a)-(b) are reproduced without modification from the arXiv version of the prior pilot study~\cite{liu2026multimodal}. Both (a) and (b) support the WoZ framework’s preliminary feasibility. (c) Response-mode and ambiguity-type distributions, revealing class imbalance and informing future communication policy design. (d) Validation (left) and test (right) confusion matrices for fine-tuned LLaVA-1.6-7B (rows: labels; columns: predictions). Fine-tuning distinguished ACT from CLARIFY and recovered ambiguity types, supporting annotation learnability.
}
\vspace{-2em}
    \label{fig: llava_finetuned}
\end{figure*}

\subsection{Feasibility of Dialogue-Driven WoZ Data Collection}
The prior pilot study reported an 80.3\% task-retention rate (53 of 66 recorded trials), multi-turn dialogue across all five tasks, and generally positive participant feedback~\cite{liu2026multimodal}. 40 episodes are retained after the dialogue analysis filtering.

\textbf{Figure~\ref{fig: llava_finetuned}(a)} demonstrates the utterances distribution among tasks, indicating the diverse with no formal communication policy. As the supplementary interactional acceptability evidence, the post-study questionnaire (\textbf{Figure~\ref{fig: llava_finetuned}(b)}) indicates that participants generally enjoyed the dialogue-driven interaction and perceived the system autonomy positively. The medians reflect strong agreement for enjoyment of the interaction and satisfaction with the autonomy, with other items tending toward agree or likely. Top-box results further support this trend with 5 assessment questions achieving 80\% top-box (ratings of 4-5) and 3 achieving 60\%.
In the open-ended responses, most participants reported enjoying using WheelArm, while one participant suggested that the robot needs to move faster. 



Overall, the retained episodes across five task categories, collecting multi-turn dialogue, and positive participant feedback support the WoZ framework’s preliminary feasibility for multimodal assistive data collection. 
\vspace{-0.5em}

\subsection{Internal Pilot Coverage and Annotation Reliability}
\vspace{-0.2em}
To evaluate if the taxonomy can cover the cases in the pilot dataset, we show the coverage analysis (Table~\ref{tab: coverage analysis}) from two human annotators. Annotator A identified 184 decision points and B identified 218 decision points. The clean rates for both human annotators were approximately 90\%, while the remaining 10\% of cases were categorized as overlapping, forced-fit, or uncovered. Analysis of the annotators’ notes showed that the two most common issues concerned the boundaries of decision-point identification and the definition of REPORT\_DONE. This informs the version 2 taxonomy of clean boundary definition on these two points.

\begin{table}[!t]
\centering
\caption{Coverage Analysis}
\label{tab: coverage analysis}
\vspace{-1em}
\begin{tabular}{|l|l|l|l|l|}
\hline
Annotator & Clean Rate & \makecell[c]{Overlap \\Rate} & \makecell[c]{Forced-fit\\ Rate} & \makecell[c]{Uncovered \\Rate}\\
\hline
Annotator A & 90.76\% &0.54\% & 8.70\% & 0\\
\hline
Annotator B & 88.99\% &9.17\% & 0 & 1.83\%\\
\hline
\end{tabular}
\vspace{-1em}
\end{table}

To prove the reproducibility of the taxonomy, two human annotators and an AI annotator independently labeled the pilot data. In IRR analysis, we use Human Annotator A's annotation as the reference, calculate Cohen's $\kappa$ and observed agreement by comparing with AI and the other human annotator. In Table~\ref{tab: Cohen}, mode end-to-end includes all items identified as decision points by either annotator, whereas mode conditional includes only items identified by both; CLARIFY detection conditional evaluates CLARIFY versus non-CLARIFY within this shared subset. Similarly, ambiguity end-to-end includes all items assigned an ambiguity label by either annotator, while ambiguity type conditional considers only items labeled CLARIFY by both annotators. The results show substantial agreement across all annotation levels for both the human–human and human–AI comparisons, as measured by Cohen’s ($\kappa$) and observed agreement.

\begin{table}[!t]
\centering
\caption{Reliability Analysis}
\label{tab: Cohen}
\vspace{-1em}
\begin{tabular}{|l|l|l|l|}
\hline
Levels  & $\kappa$ (AAI\textsuperscript{1}/AB\textsuperscript{2}) & Observed Agreement(AAI/AB) \\
\hline
Decision point & 0.86/0.92 & 0.94/0.96 \\
\hline
Mode end-to-end & 0.73/0.74 & 0.80/0.82 \\
\hline
Mode conditional & 0.95/0.86 & 0.96/0.91 \\
\hline
\makecell[l]{CLARIFY detection \\conditional} & 0.90/0.83 & 0.96/0.94 \\
\hline
Ambiguity end-to-end & 0.73/0.72 & 0.79/0.79 \\
\hline
\makecell[l]{Ambiguity type \\conditional} & 0.90/0.80 & 0.94/0.88 \\
\hline
\end{tabular}
\begin{minipage}{\textwidth}
\footnotesize
\textsuperscript{1}AAI: IRR between Human Annotator A and the AI annotator.\\
\textsuperscript{2}AB: IRR between Human Annotators A and B.\\
\end{minipage}
\vspace{-4em}
\end{table}

\begin{table*}[!t]
    \centering
    \caption{Representative robot-side communication-policy gaps observed in the pilot interactions.}
    \vspace{-1em}
    \label{tab:communication_gaps}
    \footnotesize
    \setlength{\tabcolsep}{4pt}
    \renewcommand{\arraystretch}{0.8}

    \begin{tabularx}{\textwidth}{
        @{}
        >{\raggedright\arraybackslash}X
        >{\raggedright\arraybackslash}p{0.2\textwidth}
        >{\raggedright\arraybackslash}X
        @{}
    }
        \toprule
        \textbf{Interaction Context} &
        \textbf{Observed Wizard Response} &
        \textbf{Communication-Policy Gap} \\
        \midrule

        Multiple drawers are available. &
        ``Which drawer?'' &
        Offer visible options, such as the top, middle, or bottom drawer. \\

        The user repeats the task while the robot moves slowly. &
        Repeatedly says ``Okay.'' &
        Report the current action or progress to maintain user awareness. \\

        The user requests a turn without specifying when to stop. &
        ``Okay.'' &
        Confirm the direction and clarify the stopping point. \\

        The user asks the robot to put down the drink. &
        Sometimes accepts; sometimes asks where. &
        Consistently clarify the destination when it is unspecified. \\

        Cleaning requires finding, collecting, and disposing of trash. &
        Action order varies across interactions. &
        Define a consistent policy for decomposing multi-step tasks. \\

        The user requests the ``largest apple.'' &
        ``Okay.'' &
        Clarify when the description is viewpoint-dependent or non-unique. \\

        The user requests a red apple when two are visible. &
        ``Both apples are red.'' &
        Ask a discriminating question instead of only stating the ambiguity. \\

        \bottomrule
        \vspace{-5em}
    \end{tabularx}
\end{table*}

\subsection{Communication-Policy Gaps}
We retrospectively annotated the wizard's decisions using the proposed taxonomy and used Human A's annotations as the reference for this analysis. Across 184 decision points (Figure~\ref{fig: llava_finetuned}(c)), ACT was the dominant response mode, while the other modes occurred less frequently. Among the 36 CLARIFY decisions, referential and intent-related ambiguities were most common, whereas spatial and intelligibility ambiguities were rarely observed. These distributions show that the framework elicited multiple robot-side conversational decisions, but produced imbalanced supervision that likely reflects both the selected tasks and the minimally constrained protocol. Future data collection should therefore introduce interaction contexts that deliberately elicit less frequent modes and ambiguity types.

Case-by-case analysis further revealed inconsistent wizard behavior in comparable situations (Table~\ref{tab:communication_gaps}). In particular, the wizard lacked consistent criteria for decomposing multi-step tasks, reporting progress during long-running actions, determining when information was missing, and formulating questions that distinguish among visible alternatives. These observations provide empirical requirements for refining the taxonomy and developing a more consistent wizard communication policy for future data collection.

\subsection{Feasibility of Taxonomy-Label Learnability}
With the reference annotation from human annotator A, we fine-tuned an open-source VLM model, Llava-1.6-7B~\cite{liu2024llavanext}, to test the feasibility of distinguishing modes and ambiguity with the annotation from the taxonomy. From Figure~\ref{fig: llava_finetuned}, the fine-tuned model demonstrated the ability to classify the two trained response modes, ACT and CLARIFY, as well as the associated ambiguity types.

%% file: Sections/Conclusion.tex
Overall, the prior study and recent findings support the preliminary feasibility of using a minimally constrained WoZ protocol to collect multimodal assistive dialogue and derive a practice-grounded taxonomy. They also provide preliminary evidence that the taxonomy can support reproducible annotation. By making robot decision points, response modes, and clarification-related ambiguities explicit, the taxonomy provides an initial foundation for developing more consistent wizard and robot communication policies while allowing unscripted participant responses. The next stage will refine and apply these rules in data collection with multiple teleoperators and a larger, more diverse participant sample to evaluate their coverage, consistency, and generalizability across interaction settings. Once validated, the taxonomy and a future communication policy could guide more consistent HRD data collection and provide structured supervision for developing offline conversational robot models.

%% file: reference.bib
@article{riek2012wizard,
  title={Wizard of oz studies in hri: a systematic review and new reporting guidelines},
  author={Riek, Laurel D},
  journal={Journal of Human-Robot Interaction},
  volume={1},
  number={1},
  pages={119--136},
  year={2012},
  publisher={Journal of Human-Robot Interaction Steering Committee}
}

@article{iyer2024open,
  title={Open teach: A versatile teleoperation system for robotic manipulation},
  author={Iyer, Aadhithya and Peng, Zhuoran and Dai, Yinlong and Guzey, Irmak and Haldar, Siddhant and Chintala, Soumith and Pinto, Lerrel},
  journal={arXiv preprint arXiv:2403.07870},
  year={2024}
}

@article{Ustun2003ICF,
  author    = {T. Bedirhan Üstün and Somnath Chatterji and Jerome Bickenbach and Nenad Kostanjsek and Margie Schneider},
  title     = {The International Classification of Functioning, Disability and Health: a new tool for understanding disability and health},
  journal   = {Disability and Rehabilitation},
  volume    = {25},
  number    = {11-12},
  pages     = {565--571},
  year      = {2003},
  month     = {January},
  publisher = {Taylor \& Francis},
}

@article{liu2026multimodal,
  title={A Multimodal Data Collection Framework for Dialogue-Driven Assistive Robotics to Clarify Ambiguities: A Wizard-of-Oz Pilot Study},
  author={Liu, Guangping and Hawkins, Nicholas and Madden, Billy and Sultan, Tipu and Esposito, Flavio and Babaiasl, Madi},
  journal={arXiv preprint arXiv:2601.16870},
  year={2026},
  note = {Accepted at IEEE RAS/EMBS BioRob 2026}
}

@misc{liu2024llavanext,
    title={LLaVA-NeXT: Improved reasoning, OCR, and world knowledge},
    url={https://llava-vl.github.io/blog/2024-01-30-llava-next/},
    author={Liu, Haotian and Li, Chunyuan and Li, Yuheng and Li, Bo and Zhang, Yuanhan and Shen, Sheng and Lee, Yong Jae},
    month={January},
    year={2024}
}

@article{pezzelle2023dealing,
  title={Dealing with semantic underspecification in multimodal NLP},
  author={Pezzelle, Sandro},
  journal={arXiv preprint arXiv:2306.05240},
  year={2023}
}

@article{chen2023asking,
  title={Asking before acting: Gather information in embodied decision making with language models},
  author={Chen, Xiaoyu and Zhang, Shenao and Zhang, Pushi and Zhao, Li and Chen, Jianyu},
  journal={arXiv preprint arXiv:2305.15695},
  year={2023}
}
